\documentclass[11pt,a4paper]{article}
\usepackage[hyperref]{acl2020}
\usepackage{times}
\usepackage{latexsym}
\usepackage{graphicx}
\usepackage{caption}
\usepackage{subcaption}
\usepackage{booktabs}
\usepackage{amsmath}
\usepackage{array}
\usepackage{makecell}
\usepackage{placeins}
\usepackage{tabularx}
\usepackage{adjustbox}
\usepackage{float}

\usepackage{microtype}

\aclfinalcopy 

\newtheorem{observation}{Observation}[section]

\newcommand\btrule[1]{\specialrule{#1}{2pt}{2pt}}
\newcolumntype{C}[1]{>{\centering\let\newline\\\arraybackslash\hspace{0pt}}m{#1}}

\title{Benchmarking down-scaled (not so large) pre-trained language models}

\author{Matthias~Aßenmacher\textsuperscript{$\spadesuit$} \\\And
  Patrick Schulze\textsuperscript{$\clubsuit$} \\ \\
  Department of Statistics\\
  Ludwig-Maximilians-Universität \\
  Ludwigstr. 33, D-80539 Munich, Germany \\
  \small \textsuperscript{$\spadesuit$}\texttt{\{matthias,chris\}@stat.uni-muenchen.de}, \quad \textsuperscript{$\clubsuit$}\texttt{pa.schulze@campus.lmu.de}  \\\And
  Christian Heumann\textsuperscript{$\spadesuit$}
}
\date{}

\begin{document}
\maketitle
\begin{abstract}
Large Transformer-based language models are pre-trained on corpora of varying sizes, for a different number of steps and with different batch sizes. At the same time, more fundamental components, such as the pre-training objective or architectural hyperparameters, are modified. In total, it is therefore difficult to ascribe changes in performance to specific factors. Since searching the hyperparameter space over the full systems is too costly, we pre-train down-scaled versions of several popular Transformer-based architectures on a common pre-training corpus and benchmark them on a subset of the GLUE tasks \citep{wang2018glue}. Specifically, we systematically compare three pre-training objectives for different shape parameters and model sizes, while also varying the number of pre-training steps and the batch size. In our experiments MLM + NSP (BERT-style) consistently outperforms MLM (RoBERTa-style) as well as the standard LM objective. Furthermore, we find that additional compute should be mainly allocated to an increased model size, while training for more steps is inefficient. Based on these observations, as a final step we attempt to scale up several systems using compound scaling \citep{tan2019efficientnet} adapted to Transformer-based language models.
\end{abstract}

\section{Introduction}
\label{sec:intro}

The introduction of the Transformer \citep{vaswani2017attention} together with the application of transfer learning \citep{thrun1998learning} has led to major advances in Natural Language Processing (NLP) in recent years. While many different lines of research exist, most attention is generally paid to the largest systems which often reach new state-of-the-art (SOTA) results. The current trend is to scale up such systems to ever new orders of magnitude: 213M parameters in the original Transformer \citep{vaswani2017attention}, 300M parameters in BERT \citep{devlin2019bert}, 1.5B parameters in GPT-2 \citep{radford2019language} and recently 175B parameters in GPT-3 \citep{brown2020language}. Since these models are pre-trained on corpora of widely varying sizes, for a different number of training steps and with different batch sizes, comparability suffers \citep{assenmacher2020comparability}. At the same time, new systems often apply fundamentally different methods, such as using a different pre-training objective or modified architectural hyperparameters. While altering multiple components simultaneously can help achieve new SOTA results, which is an important endeavor, it is difficult to disentangle the effects of the various factors. Though there exist various ablation studies, these often show only a small excerpt from the broad spectrum of experimental opportunities and are thus not suitable for providing a comprehensive picture. 

\section{Related work}
\label{sec:related}

One line of research empirically derives generalization results for large neural NLP systems. \citet{rosenfeld2019constructive} study how the generalization error of language models (LMs) depends on model and data set size. Regarding model size, they provide an approximation of the test loss, assuming that a LM is scaled with respect to a pre-defined scheme, such as increasing solely the embedding dimension. A related but more comprehensive study was conducted by \citet{kaplan2020scaling}, examining power laws of the test loss when scaling large neural LMs with respect to a broad variety of different dimensions. These dimensions include architectural hyperparameters, model size, data set size, number of training steps and batch size. A central question in their work is how these factors can be combined to attain an optimal performance given a fixed amount of compute. 

Compute efficient training is also investigated by \citet{li2020train}, recognizing that an optimal allocation of computational resources is crucial for improving model performance. Considering Masked Language Modeling (MLM) pre-training, \citet{li2020train} examine the optimal choice of number of training steps and batch size in the relation to the model size. In a large-scale study, \citet{raffel2019exploring} cover an even broader variety of modeling scenarios than \citet{kaplan2020scaling}, but train a much smaller number of systems per scenario. For instance, they include several variants of the Transformer, different pre-training objectives and various fine-tuning strategies in their analysis. Finally, based on their observations, \citet{raffel2019exploring} also scale-up a system to $11$B parameters.

\section{Materials and Methods}
\label{sec:models}

\paragraph{Pre-training data} We pre-train all models on WikiText-103\footnote{\href{https://www.salesforce.com/products/einstein/ai-research/the-wikitext-dependency-language-modeling-dataset/}{www.salesforce.com/products/einstein/ai-research/the-wikitext-dependency-language-modeling-dataset/}} \citep{merity2016pointer}, a large-scale text corpus for training and evaluating language models on long-range contexts, which has served as an evaluation data set \citep{radford2019language, dai2019transformer, shoeybi2019megatron} as well as for pre-training \citep{howard2018universal}. We pre-train all models on the training set of WikiText-103, which allows for learning long-range dependencies \citep{rae2019compressive}. The validation set is employed to compare different architectures by their validation loss during pre-training. WikiText-103 is much smaller than most pre-training corpora of modern language models. For instance, \citet{devlin2019bert} trained BERT on a $3,300$M words corpus, which is approximately 32x the size of WikiText-103. Aside from this, pre-training data sets of different models often vary considerably in size, which makes fair comparisons difficult \citep{assenmacher2020comparability}. Pre-training on the same corpus allows us to exclude the amount and quality of pre-training data as confounding factors when evaluating the different model components.

\paragraph{Models} We compare three different model types: BERT \citep{devlin2019bert}, RoBERTa \citep{liu2019roberta} and GPT-2 \citep{radford2019language}. BERT is a bidirectional Transformer encoder which is trained with both MLM as well as Next Sentence Prediction (NSP). Its direct successor RoBERTa relies on the exact same architecture and differs from BERT solely in the pre-training procedure. Amongst other changes, \citet{liu2019roberta} abandoned the NSP objective and introduced a dynamic masking\footnote{We also use dynamic masking throughout this study.} procedure for the MLM objective\footnote{There were further alterations, none of which are crucial for our experiments since we are using fixed pre-training data sets, batch sizes, learning rates, etc. for better comparability.}. GPT-2 is a Transformer decoder, and thus a unidirectional model, trained with the standard LM objective.

Since we train a multitude of down-scaled versions for each model type, thus modifying the specifications of the original models, we introduce the following conventions: We label models trained with MLM \& NSP as \textit{BERT-style}, models trained with MLM as \textit{RoBERTa-style}, and models trained with LM as \textit{GPT-2-style}. Alongside with the pre-training objectives, we also use the respective tokenizers of the different models. 

\paragraph{Fine-tuning data} We fine-tune and evaluate our systems on GLUE \citep{wang2018glue}. We mainly compare performances on MNLI \citep{williams2017broad}, QQP \citep{quora} and QNLI \citep{wang2018glue}, which are the three largest GLUE tasks, since the results on these tasks are the most reliable. In particular, we therefore calculate the average score over the validation set performances of the three tasks, which we denote by \textit{GLUE-Large}. For MNLI, we consider only the matched validation set when calculating this score. Whenever meaningful results for the two next largest data sets SST-2 \citep{socher2013recursive} and CoLa \citep{warstadt2019neural} were achieved, those will also be reported.

\paragraph{Training details} Hyperparameters and the pre-training/fine-tuning procedure are largely adopted from the original models (cf. Appendix \ref{a:pretrain} and \ref{a:finetune}).

\section{Experiments}
\label{sec:experiments}

\subsection{Comparison of different Shapes\footnote{There exist several other choices, but examining the entire spectrum of possible shapes is out of the scope of this study.}}
\label{sec:shapes}

In computer vision it has been observed that the performance of a neural network strongly depends on the choice of architectural hyperparameters, such as width or depth \citep{tan2019efficientnet}. In contrast, \citet{kaplan2020scaling} observed a similar LM test loss over a wide range of shape parameters. Similarly, for MLMs, \citet{li2020train} found that the validation loss does not depend strongly on the model shape. This holds true also for the MNLI validation accuracy of fine-tuned systems.

In this study, we examine the impact of three different architectural hyperparameters in Transformer-based models: \textit{depth}, \textit{width} and the \textit{number of attention heads}. Depth is given by the number of layers $L$. Stacking many layers in Transformer-based systems can be somewhat inefficient and does not always lead to a considerable increase in performance \citep{lan2019albert}. Width corresponds to the embedding dimension $H$. Increasing $H$ has in general produced slightly better results than increasing $L$ in Transformer-based systems \citep{lan2019albert, raffel2019exploring, li2020train}. Attention Heads are used to discriminate between different regions of the embedding space. In most applications of the Transformer, the number of attention heads $A$ is set in fixed relation to $H$, such as $H = 64 \times A$. Decreasing performance has been reported for larger ratios \citep{vaswani2017attention, brown2020language}.

\subsection{Model Size, Training Steps and Batch size}
\label{sec:size}

Several recent studies have investigated the problem of compute efficient training of Transformer-based systems \citep{raffel2019exploring, li2020train, kaplan2020scaling}. The consensus among these studies is that, under a restricted budget, optimal performance is achieved by training very large models and stopping training well before convergence. Furthermore, additional compute should rather be used to increase the batch size instead of training for more steps. To examine convergence characteristics, we monitor the pre-training validation loss of several systems and test how this loss corresponds to different model sizes and shapes. Additionally, we conduct experiments regarding the effect of the batch size and the number of training steps. In particular, we evaluate how the training time and the model performance depend on both factors.

\subsection{Definition of the Model Size}
\label{sec:params}

We follow \citet{kaplan2020scaling} and use the approximate number of non-embedding parameters to define the model size, which we denote as $N_{\mbox{model}}$. Since the share of embedding parameters decreases for larger models, similarly to \citet{kaplan2020scaling} we expect that discarding the number of embedding parameters allows for better generalization of our results to large models. Another advantage of defining the model size as the number of non-embedding parameters is that it is closely linked to the number of (non-embedding related) FLOPs \citep{kaplan2020scaling}. This enables us to design benchmarking scenarios by training different models of comparable size, which at the same time require roughly similar amounts of computation.

Omitting biases and other sub-leading terms, the number of non-embedding parameters is given by
\begin{equation}
N_{\mbox{model}} = 12 L H^2,
\label{eq:nmod}
\end{equation}
assuming that queries, keys and values are all transformed to dimension $\frac{H}{A}$ and the feed-forward dimension is $4H$. For a more in-depth explanation, please see Appendix \ref{a:modelsize}. 

\section{Results\footnote{Source code: \href{https://github.com/PMSchulze/NLP-benchmarking}{\scriptsize https://github.com/PMSchulze/NLP-benchmarking}}
}
\label{sec:results}

We start by evaluating how varying single shape dimensions affects the performance on GLUE-Large for the three different pre-training objectives (cf. Sec. \ref{sec:single}). This aims as investigating whether the performance gain diminishes after a certain level, comparing how the performance changes when scaling different dimensions, and examining whether models with different pre-training objectives respond differently to single-dimension scaling. Subsequently in Section \ref{sec:multi}, we change multiple shape dimensions simultaneously to investigate whether the different dimensions depend on each other. In Sections \ref{sec:loss} and \ref{sec:batchsize_steps} we study how to train efficiently by varying the model size, the number of training steps and the batch size. In Section \ref{sec:scale} we put together our observations from the previous sections and scale networks to different sizes.

\begin{figure*}[t!]
  \centering
  \captionsetup{margin=0cm}
  \begin{subfigure}[b]{0.49\linewidth}
    \includegraphics[width=\linewidth]{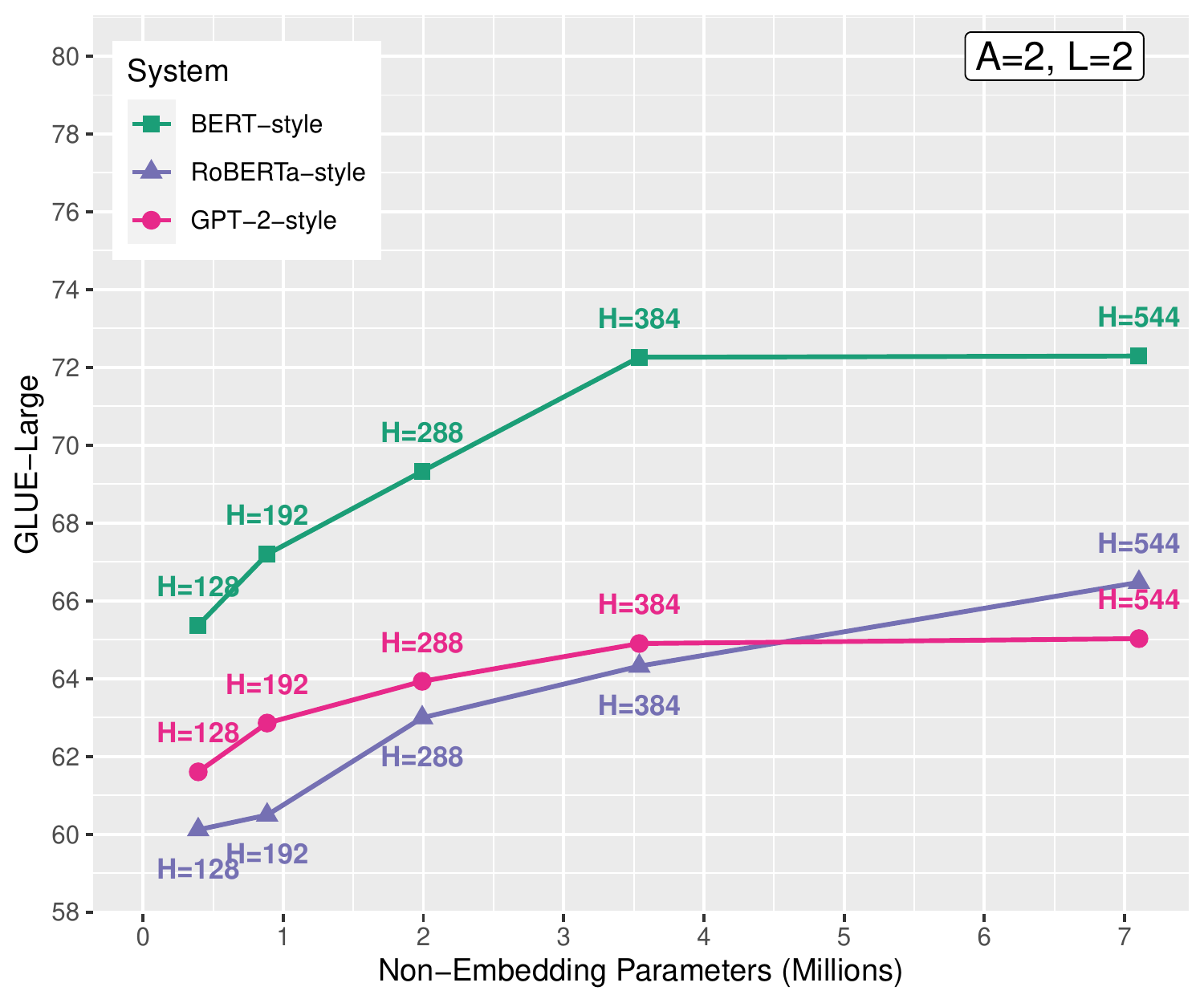}
  \end{subfigure}
  \begin{subfigure}[b]{0.49\linewidth}
    \includegraphics[width=\linewidth]{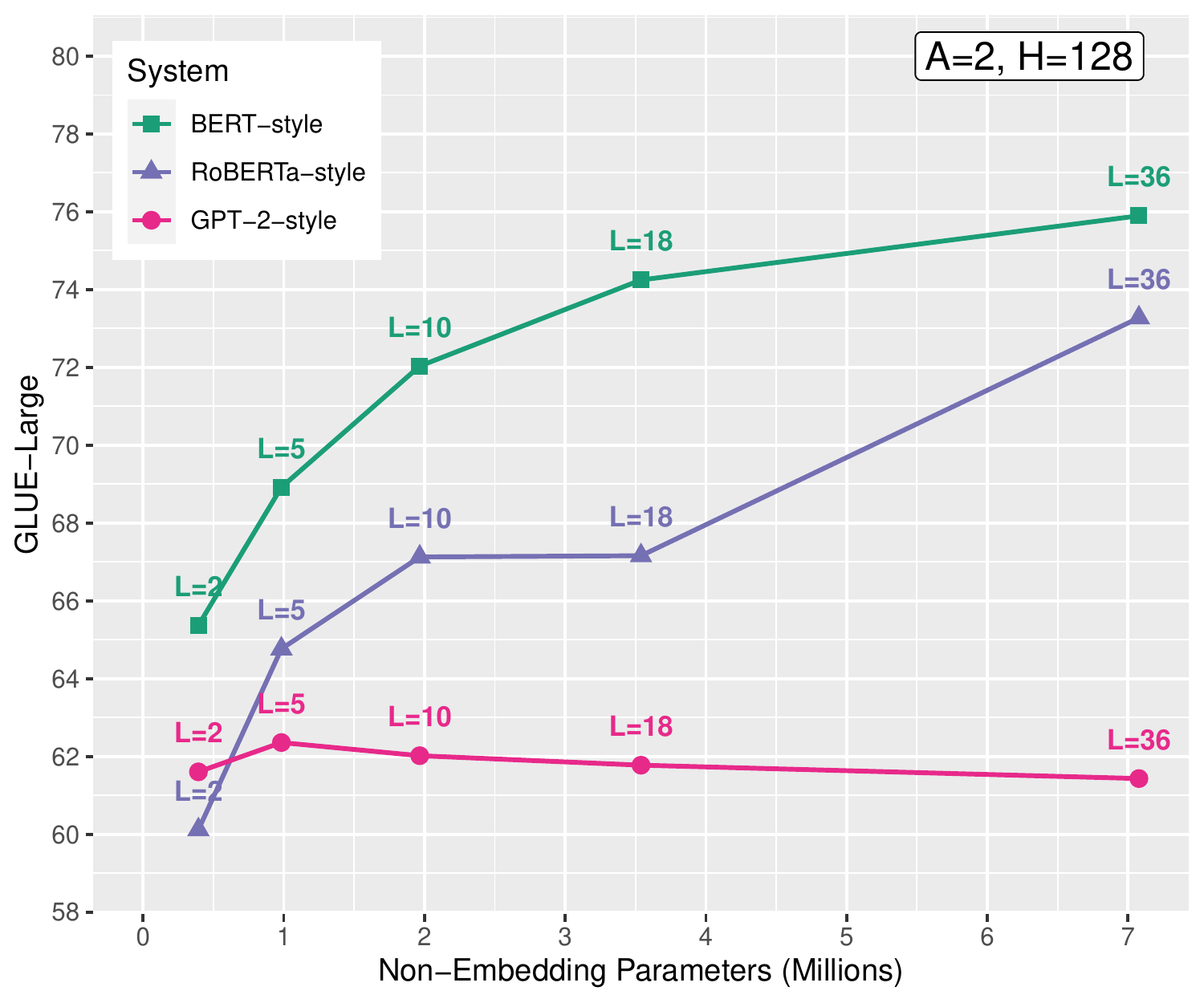}
  \end{subfigure}\\
  \caption{Average score on GLUE-Large, when varying $H$ (left) vs.\ when varying $L$ (right). For detailed performance values on the single tasks, see Table \ref{tab:embed} and Table \ref{tab:layer} in Appendix \ref{a:performance}.}
  \label{fig:single_dimension}
\end{figure*}

\subsection{Scaling Single Shape Dimensions} \label{sec:single}

In this section, we separately scale $L$ and $H$, while holding all other dimensions constant. As shown in Figure \ref{fig:single_dimension}, BERT-style systems perform significantly better than GPT-2-style and RoBERTa-style systems on GLUE-Large, contrary to the results of \citet{liu2019roberta} and in line with the original findings of \citet{devlin2019bert}.

\begin{observation}
The pre-training objective has a large impact on the performance of a fine-tuned system. Pre-training with the combination of MLM \& NSP achieves the best results on sentence-pair tasks, while pre-training with the unidirectional LM objective shows in general the worst performance.
\end{observation}

\noindent Furthermore, for BERT-style systems the average performance is a relatively smooth function of the model size. Scaling up either the $L$ or $H$ results in an initial increase in performance, which then saturates at approximately $75\%$ and $72\%$, respectively. For RoBERTa-style systems, the difference between scaling $L$ and increasing $H$ is much larger. Furthermore, although the performance seems to saturate at a certain level, the trend is less apparent than for BERT-style systems.\footnote{Note that the relatively low average score for the 18-layer RoBERTa-style system, shown in the right plot of Figure \ref{fig:single_dimension}, is due to a weak performance on the QNLI task.} For GPT-2-style systems, the average score slightly increases when scaling the embedding size, but interestingly, stacking more layers shows no positive effect at all. This suggests that GPT-2-style systems require more pre-training data compared to BERT-style and RoBERTa-style systems.

\begin{observation}
In most cases, the performance of a fine-tuned system increases up to a certain level when scaling either width or depth, but the progression depends strongly on the pre-training objective. 
\end{observation}

\subsection{Scaling Multiple Shape Dimensions} \label{sec:multi}

\begin{figure*}[t!]
  \centering
  \captionsetup{margin=0cm}
  \begin{subfigure}[b]{0.49\linewidth}
    \includegraphics[width=\linewidth]{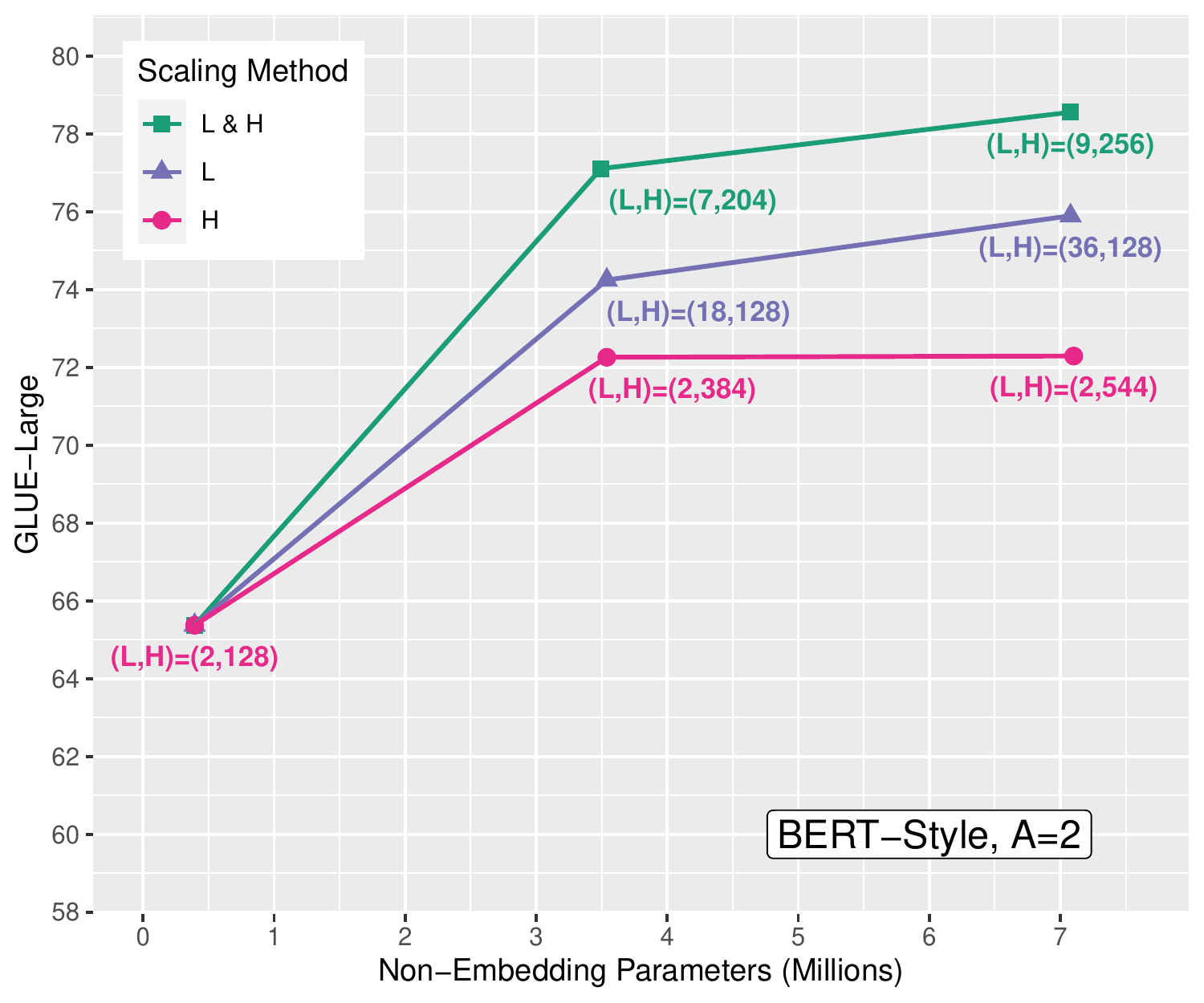}
  \end{subfigure}
  \begin{subfigure}[b]{0.49\linewidth}
    \includegraphics[width=\linewidth]{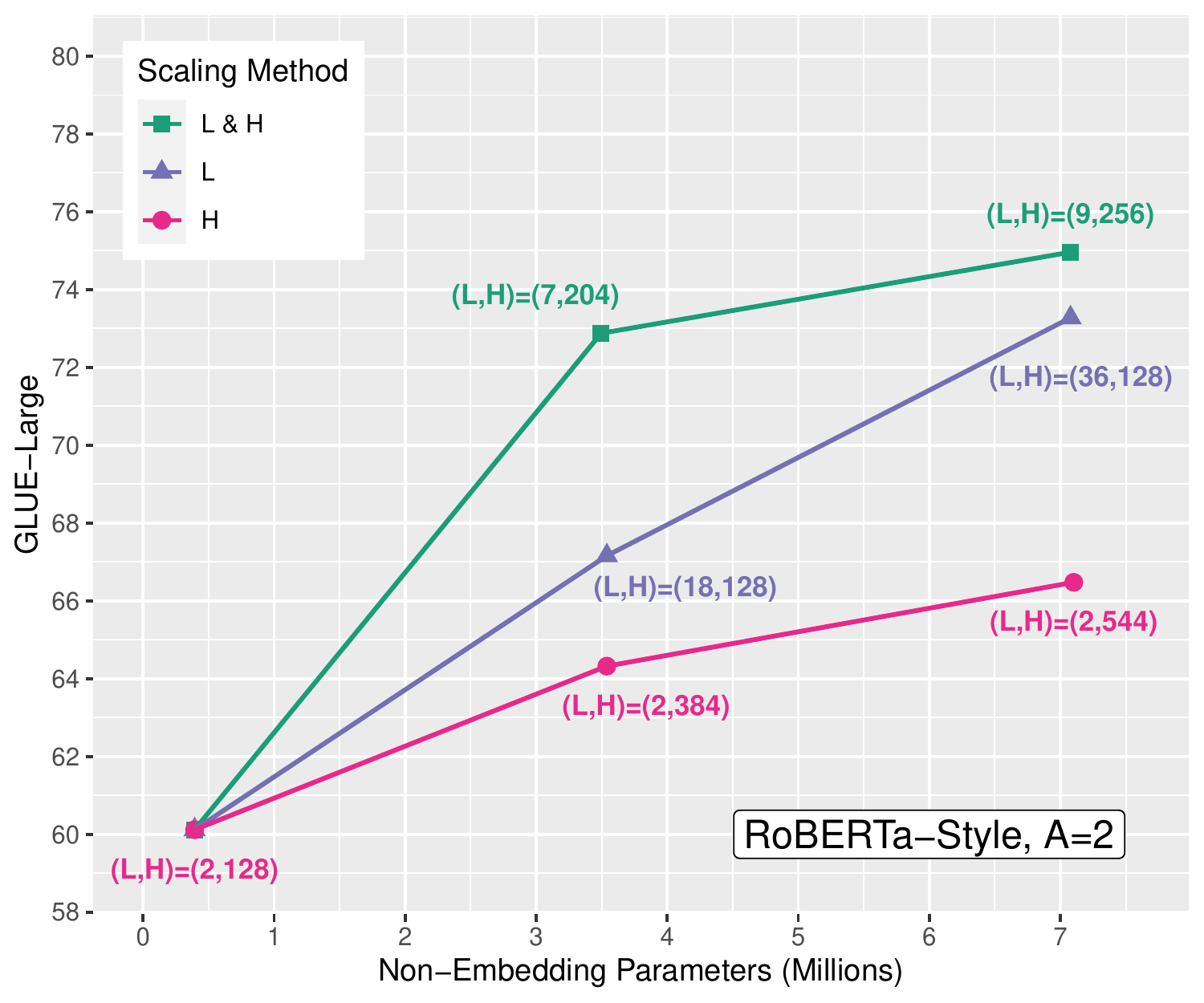}
  \end{subfigure}\\
  \caption{Performance on GLUE-Large when increasing multiple shape dimensions.}
  \label{fig:comparison_single_mutli}
\end{figure*}

We next examine whether the performance can be improved by scaling multiple dimensions at the same time. First, we increase both $H$ and $L$ and compare the performance with the results from Section \ref{sec:single}. Fig. \ref{fig:comparison_single_mutli} shows that for RoBERTa-style and BERT-style systems, scaling both dimensions significantly improves the performance on GLUE-Large.
\begin{observation}
Scaling multiple shape dimensions can lead to a better performance than scaling single dimensions.
\end{observation}
\noindent Therefore, we conclude that the shape dimensions are not independent of each other. For GPT-2-style systems, however, we do not observe a performance increase, as shown in Table \ref{tab:multi_dimension}.

\begin{table}[ht]
\small
\renewcommand{\arraystretch}{0.8}
\begin{tabular*}{\linewidth}{ @{\extracolsep{\fill}} ccccc}
\btrule{1.5pt}
\multicolumn{4}{c}{\textbf{BERT-Style}} &
\textbf{Validation Set Performance} \\
\cmidrule(r){1-4}
\cmidrule(r){5-5}
A & H & L & $N_{\mbox{model}}$ & $\mbox{GLUE-Large}$  \\
\midrule
2 & 204 & 7 & 3,495,744 & 77.1 \\
2 & 256 & 9 & 7,077,888 & 78.6 \\
8 & 544 & 2 & 7,102,464 & 78.4 \\
\midrule
\multicolumn{4}{c}{\textbf{GPT-2-Style}} &
\textbf{Validation Set Performance} \\
\cmidrule(r){1-4}
\cmidrule(r){5-5}
A & H & L & $N_{\mbox{model}}$ & $\mbox{GLUE-Large}$ \\
\midrule 
2 & 204 & 7 & 3,495,744 & 63.6 \\
2 & 256 & 9 & 7,077,888 & 63.8 \\
8 & 544 & 2 & 7,102,464 & 66.0 \\
\midrule
\multicolumn{4}{c}{\textbf{RoBERTa-Style}} &
\textbf{Validation Set Performance} \\
\cmidrule(r){1-4}
\cmidrule(r){5-5}
A & H & L & $N_{\mbox{model}}$ & $\mbox{GLUE-Large}$ \\
\midrule
2 & 204 & 7 & 3,495,744 & 72.9 \\
2 & 256 & 9 & 7,077,888 & 75.0 \\
8 & 544 & 2 & 7,102,464 & 70.9 \\
\btrule{1.5pt}
\end{tabular*}
\caption{Performance on GLUE-Large when increasing multiple shape dimensions at the same time.}
\label{tab:multi_dimension}
\end{table}

So far, we did not increase $A$ when scaling $H$ and observed that, without using more attention heads, wide systems perform worse than deep systems (cf. Fig. \ref{fig:single_dimension}). To evaluate whether a larger number of attention heads can boost the performance of wide systems, we re-implement our widest systems with $A=8$ attention heads, which corresponds to $\frac{H}{A} = 68$. We observe that the score of the widest system on GLUE-Large improved substantially by doing so (cf. Fig. \ref{fig:single_dimension} and Tab. \ref{tab:multi_dimension}). In particular, when using $A = 8$ instead of $A = 2$, the wide BERT-style system ($A=8, H=544, L=2$) performs even better than the deep BERT-style system of comparable size ($A=2, H=128, L=36$). Furthermore, as also shown in Table \ref{tab:multi_dimension}, the wide BERT-style system (with increased $A$) performs close to the balanced one ($A=2$, $H=256$, $L=9$).
\begin{observation}
The fine-tuning performance can be similar over a wide range of shapes. For BERT-style systems, wide systems perform slightly better than deep systems, if the number of attention heads is adapted to the embedding dimension.
\label{shape_weak}
\end{observation}
\noindent In contrast to BERT-style systems, deep RoBERTa-style systems still perform better than wide systems, even when increasing the number of attentions heads. For GPT-2-style systems, adding more attention heads hardly increases the performance.

\subsection{Monitoring the Validation Loss} \label{sec:loss}

In the previous sections, different models were made comparable by their number of non-embedding parameters. As stated in section \ref{sec:params}, this number is related to the computational cost when evaluated as the number of FLOPs. Reporting the computational cost in FLOPs neglects, however, that some operations can be run in parallel, while others cannot. In order to assess the speed of convergence, following \citet{li2020train}, we therefore directly report the wall-clock time in seconds.

\begin{figure*}[t!]
  \centering
  \captionsetup{margin=0cm}
  \begin{subfigure}[b]{0.49\linewidth}
    \includegraphics[width=\linewidth]{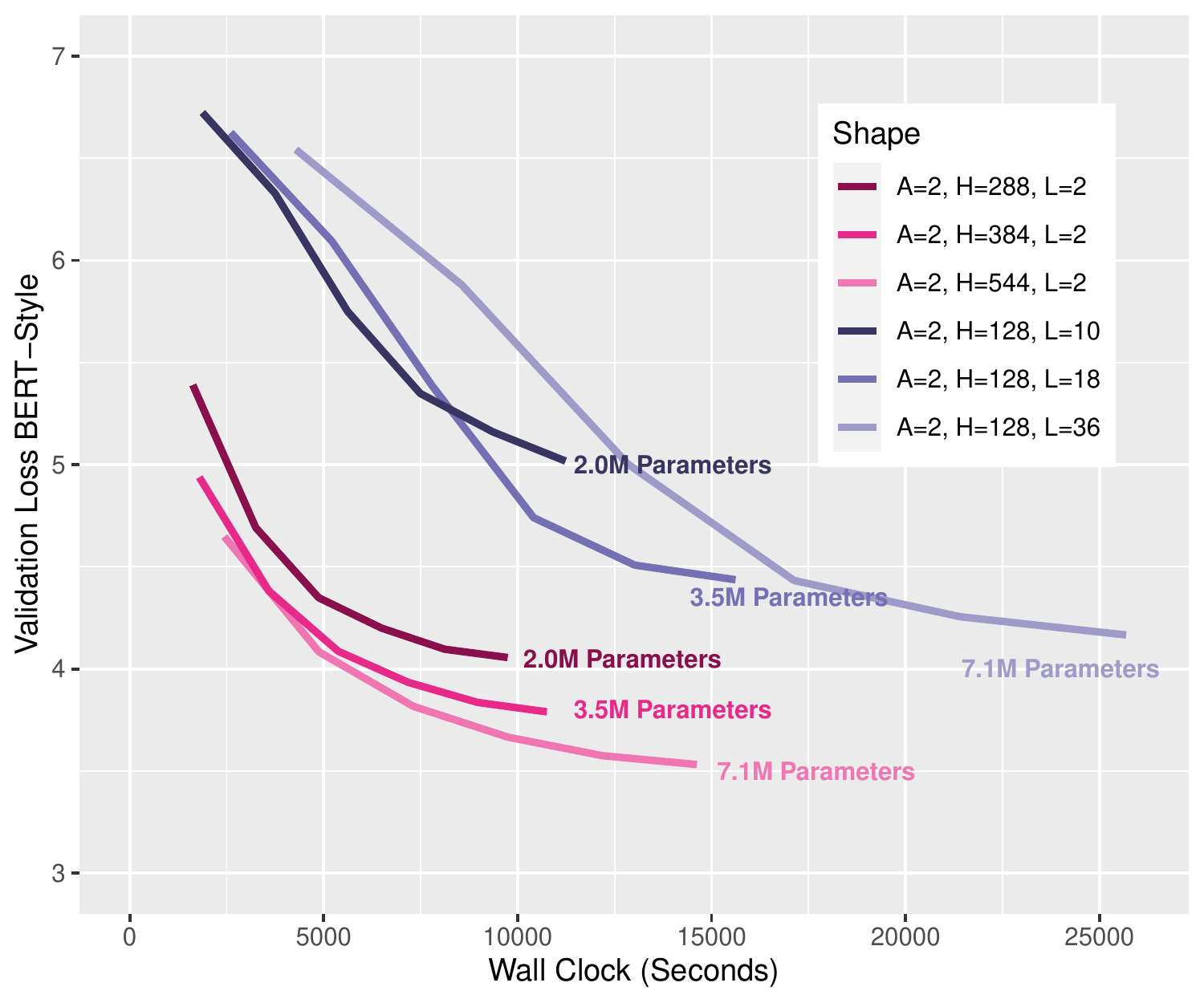}
  \end{subfigure}
  \begin{subfigure}[b]{0.49\linewidth}
    \includegraphics[width=\linewidth]{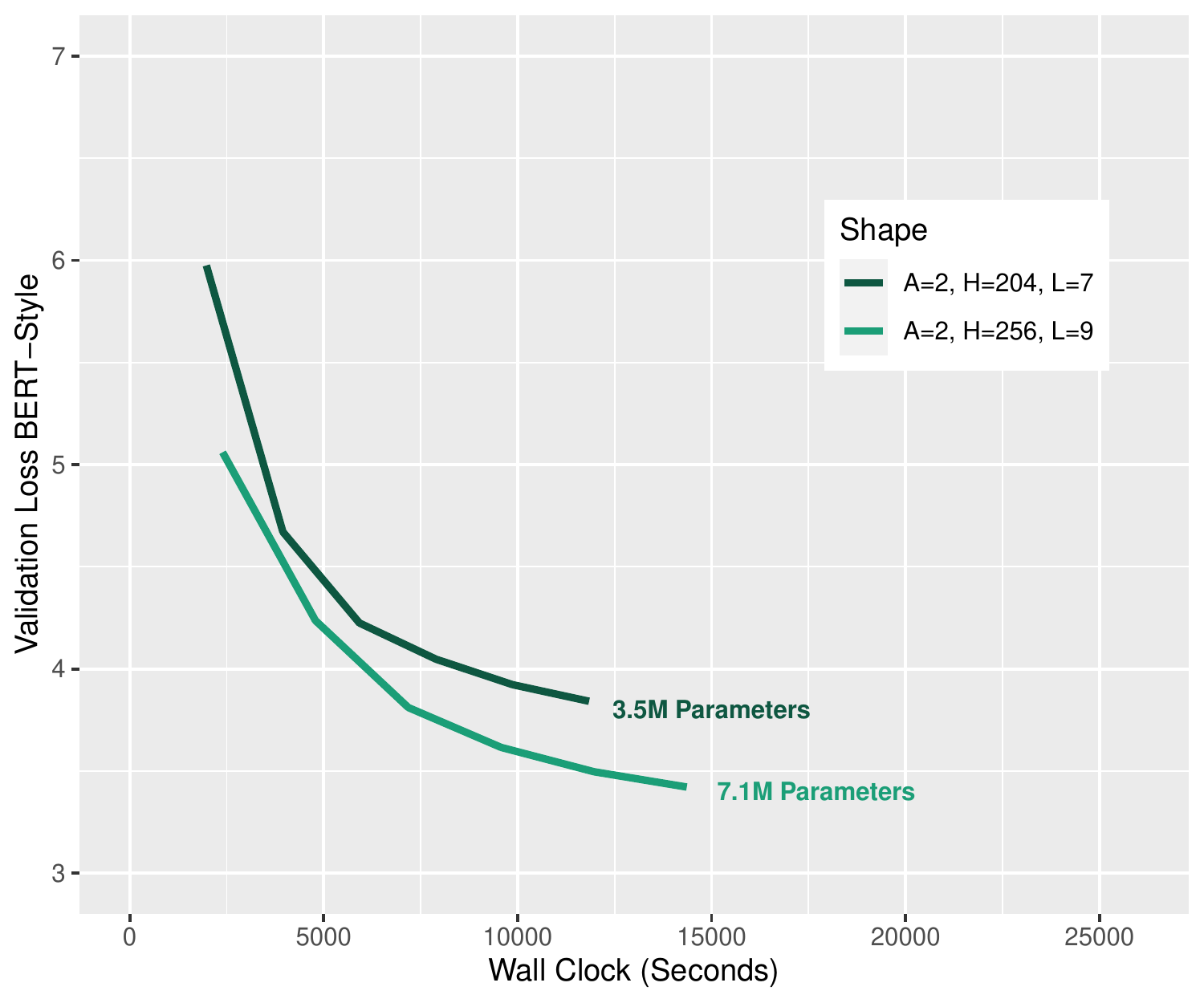}
  \end{subfigure}\\
  \caption{Loss curves of BERT-Style systems of different shape. All loss curves are associated with the first stage of pre-training, where we train on short sequences with a of 128 tokens (For the loss curves for the subsequent training on the long sequences, see Appendix \ref{a:val_loss}). The depicted parameter counts refer to the model size $N_{model}$.}
  \label{fig:loss_bert_short}
\end{figure*}

Figure \ref{fig:loss_bert_short} shows the validation loss for BERT-style systems of different shape, when pre-trained on the short sequences.\footnote{We do pre-training on short and long sequences. For a detailed description, see Appendix \ref{a:pretrain} and Appendix \ref{a:input}.}. The left plot depicts several pre-training loss curves corresponding to the single-dimension scaling experiments from Section \ref{sec:single}. Interestingly, when comparing the validation loss with the GLUE-Large results (cf. Fig. \ref{fig:single_dimension}), we find that, although increasing $H$ (while holding $A$ fixed) results in a lower validation loss than increasing $L$, the GLUE-Large score shows a higher increase in the latter case.
\begin{observation}
The pre-training validation loss is not necessarily a good indicator for the performance of a fine-tuned system.
\label{loss_indication}
\end{observation}
\noindent Dependent on the downstream task some architectures presumably favor fine-tuning more than others, which can offset a relatively worse initialization point. This finding suggests that, although \citet{kaplan2020scaling} observe similar test losses for different shapes, benchmarking the corresponding fine-tuned versions may present a different picture.

In the left plot of Figure \ref{fig:loss_bert_short} we furthermore observe that shape has a significant effect on the pre-training time. In particular, stacking many layers requires much longer pre-training. It is also evident that increasing the size does not lead to a proportionate increase in the pre-training time. This holds true especially when scaling multiple dimensions, as depicted in the right plot of Figure \ref{fig:loss_bert_short}. When doubling the number of pre-training parameters, the training time only increases from approximately $11,800$ seconds to approximately $14,400$ seconds. In particular, the loss of the larger system is smaller at any measured point in time.
\begin{observation}
Given a fixed time budget, training large systems for a relatively small number of steps is more efficient than training small systems for a large number of steps.
\label{train_large}
\end{observation}
\noindent The 9-layer system in the right plot of Figure \ref{fig:loss_bert_short} achieves a notably lower validation loss than the 7-layer system after $10,000$ seconds, which corresponds to approximately $65,800$ and $79,800$ steps, respectively. \citet{li2020train} made a similar observation by showing that larger Transformer-based systems generally reach a lower pre-training validation perplexity in shorter time. A point of concern might be that larger systems overfit more easily during fine-tuning. However, \citet{li2020train} showed that, when stopping models of different size at the same pre-training validation perplexity, large systems generally achieve comparable downstream task performances to small systems, which contradicts the overfitting argument.

\subsection{Number of Training Steps and Batch Size} \label{sec:batchsize_steps}

The amount of processed data can be increased by increasing either the number of trainig steps or the batch size. In Table \ref{batchsize_training_steps} we compare how halving the number of steps vs.\ halving the batch size impacts model performance. As baseline we use our best performing system thus far ($A=2, H=256, L=9$), pre-trained RoBERTa- and BERT-style.

\begin{table*}[t!]
\footnotesize
\renewcommand{\arraystretch}{0.8}
\captionsetup{margin=0cm}
\begin{tabular*}{\textwidth}{ @{\extracolsep{\fill}} ccccccc}
\btrule{1.5pt}
\multicolumn{2}{c}{\textbf{BERT-Style}} &
\multicolumn{5}{c}{\textbf{Validation Set Performance}} \\
\cmidrule(r){1-2}
\cmidrule(r){3-7}
Training Strategy & Total Time & $\mbox{GLUE-Large}$ & MNLI-(m/mm) & QQP &  QNLI & SST-2 \\
\midrule
Baseline & $21,358$s & 78.6 & $72.0/72.7$ & $81.2$ & $82.5$ & $83.4$  \\
$\frac{1}{2}$x steps, $1$x batch & $10,736$s & 77.4 & $70.2/71.2$ & $80.5$ & $81.5$ & $82.5$  \\
$1$x steps, $\frac{1}{2}$x batch & $14,575$s & 78.2 & $71.5/71.9$ & $80.9$ & $82.3$ & $83.9$ \\
\midrule
\multicolumn{2}{c}{\textbf{RoBERTa-Style}} &
\multicolumn{5}{c}{\textbf{Validation Set Performance}} \\
\cmidrule(r){1-2}
\cmidrule(r){3-7}
Training Strategy & Total Time & $\mbox{GLUE-Large}$ & MNLI-(m/mm) & QQP &  QNLI & SST-2  \\
\midrule
Baseline & $19,760$s & 75.0 & $68.4/70.9$ & $78.2$ & $78.3$ & $75.0$  \\
$\frac{1}{2}$x steps, $1$x batch & $9,906$s & 73.7 & $67.0/69.0$ & $76.7$ & $77.4$ & $83.5$   \\
$1$x steps, $\frac{1}{2}$x batch & $13,101$s & 75.6 & $68.2/70.0$ & $79.5$ & $78.9$ & $84.4$  \\
\btrule{1.5pt}
\end{tabular*}
\caption{GLUE results and total pre-training time when halving batch size vs. number of training steps.}
\label{batchsize_training_steps}
\end{table*}

In both cases we find that reducing the number of training steps is more detrimental to the performance than reducing the batch size. Conversely, it follows that when scaling up a system, a better model performance can be achieved when doubling the amount of training steps than when doubling the batch size, which is consistent with the results of \citet{raffel2019exploring}. On the other hand, we observe that the systems with the smaller batch size were trained for a significantly longer time than the systems with the reduced number of training steps. Therefore, increasing the batch size may result in a more favorable training duration than increasing the number of training steps. The modest drop in GLUE-Large performance, when halving the number of training steps is consistent with our findings from Section \ref{sec:loss} and provides additional evidence that training for a large number of steps is inefficient. 
\begin{observation}
Doubling the number of training steps marginally increases the downstream task performance, whereas doubling the batch size significantly reduces the average training time of an input sequence.
\end{observation}
\noindent As stated, several other studies have shown that using a larger batch size is in general more efficient than training for more steps \cite{kaplan2020scaling}. This means that the reduction of training time by using larger batches dominates the marginal performance gains resulting from an increased number of training steps. However, for each specific model and training configuration there exists a critical batch size, after which the performance hardly improves, if at all \cite{kaplan2020scaling, li2020train}. Our results suggest that this critical size is very small in our experiments, which we believe is due to the small size of the pre-training data set, as also observed by \citet{kaplan2020scaling}.

\subsection{Systematic Scaling}\label{sec:scale}

In this section we apply a modified version of the compound scaling method that was used to scale up EfficientNet \cite{tan2019efficientnet}, achieving a notably better accuracy on ImageNet \cite{deng2009imagenet} than previous approaches using less compute. For scaling, we only consider BERT-style systems and propose the following compound scaling method for Transformer-based systems:
\begin{align}
\begin{split}
& L = \alpha^{\phi}, \qquad H = \beta^{\phi}, \qquad A \approx H/64, \\
& \text{s.t. } \alpha \beta^2 \approx 2, \quad \mbox{with} \quad \alpha \geq 1, \beta \geq 1.\label{compound}
\end{split}
\end{align}
For suitable values of $\alpha$ and $\beta$, a system is scaled up by increasing the \textit{compound coefficient} $\phi$. Doubling $L$ doubles $N_{\text{model}}$, while $H$ leads to a fourfold increase. Since $N_{\text{model}}$ dominates the amount of compute in a Transformer, the constraint $\alpha \beta^2 \approx 2$ thus ensures that when scaling the network from $\phi_{\text{old}}$ to $\phi_{\text{new}}$, the amount of compute (which is approx. independent of $A$) approximately increases by the factor $2^{\phi_{\text{new}}-\phi_{\text{old}}}$. Following existing approaches and using Observation \ref{shape_weak}, we therefore set the number of attention heads to $A \approx H/64$.

\begin{table*}[t!]
\small
\renewcommand{\arraystretch}{0.8}
\begin{tabular*}{\textwidth}{ @{\extracolsep{\fill}} ccccccccc}
\btrule{1.5pt}
\multicolumn{7}{c}{\textbf{BERT-Style}} &
\multicolumn{2}{c}{\textbf{Validation Set Performance}} \\
\cmidrule(r){1-7}
\cmidrule(r){8-9}
$\phi$ & A & H & L & $N_{\text{model}}$ & Total Time & Epochs & $\text{GLUE-Large}$ & Final Loss  \\
\midrule
\texttt{NA} & 2 & 256 & 9 & 7,0778,88 & 21,358s & 6 & 78.6 & 3.24 \\
\texttt{NA} & 4 & 256 & 9 & 7,0778,88 & 21,703s & 6 & 78.9 & 3.29 \\
19.865 & 7 & 469 & 4 & 10,558,128 & \textbf{20,873s} & 5 & \textbf{79.4} & \textbf{3.13} \\
\btrule{1.5pt}
\end{tabular*}
\caption{Verification of the scaling method: The proposed modifications lead to a better GLUE score and a lower validation loss, while requiring less training time compared to previous best performing models.}
\label{tab:verification}
\end{table*}

\begin{table*}[t!]
\small
\renewcommand{\arraystretch}{0.8}
\begin{tabular*}{\textwidth}{ @{\extracolsep{\fill}} ccccccccccc}
\btrule{1.5pt}
\multicolumn{5}{c}{\textbf{BERT-Style}} &
\multicolumn{6}{c}{\textbf{Validation Set Performance}} \\
\cmidrule(r){1-5}
\cmidrule(r){6-11}
$\phi$ & A & H & L & $N_{\mbox{model}}$ & $\mbox{GLUE-Large}$ & MNLI-(m/mm) & QQP &  QNLI & SST-2 & CoLA  \\
\midrule
20.578 & 9 & 585 & 5 & 20,553,500 & 80.7 & $75.3/75.5$ & $83.5$ & $83.4$ & $85.1$ & $16.5$ \\
21.716 & 13 & 832 & 5 & 41,553,440 & \textbf{81.4} & \textbf{75.6/75.9} & \textbf{84.1} & \textbf{84.4} & \textbf{85.8} & \textbf{21.3} \\
\btrule{1.5pt}
\end{tabular*}
\caption[GLUE results: scaling to large sizes]{GLUE results of BERT-style systems, scaled up based on the observations made in the previous sections.}
\label{tab:final_scaling}
\end{table*}

\paragraph{Grid search} To determine $\alpha$ and $\beta$, we follow \citet{tan2019efficientnet} and perform a grid search over a set of nine small networks of comparable size trained only on the short sequences. Subsequently, we select the three systems with the lowest validation loss. Based on Observation \ref{loss_indication}, we then fine-tune and evaluate these three systems on GLUE-Large, which leads to the best performing system having $L=3$ and $H=104$ (cf. Tab. \ref{tab:grid_search} in Appendix \ref{a:performance}). From the constraint in Eq.\ \eqref{compound} it follows that the size of this system corresponds to a compound coefficient of $\phi = \log_2 (LH^2) = 14.99 \approx 15$, such that we obtain $\alpha = 3^{\frac{1}{15}} \approx 1.076$, $\beta = 104^{\frac{1}{15}} \approx 1.363$. Note that the resulting coefficients favor scaling width over depth. In general, we believe that this is reasonable, especially in light of the much longer training times of deep networks compared to wide networks (cf. Fig. \ref{fig:loss_bert_short}). However, we also want to emphasize that further research is needed, whether these scaling coefficients are suitable for BERT-style systems. For GPT-2-style systems, \citet{kaplan2020scaling} proposed to scale such that width/depth remains fixed. Importantly, however, \citet{kaplan2020scaling} did not study the effect of shape parameters on the GLUE-Large performance, but instead only monitored the LM test loss. In machine translation, on the other hand, Transformer-based systems are scaled preferably by increasing width \cite{shazeer2018mesh, li2020train}. Other approaches focus on increasing depth, while making modifications to the Transformer to allow for more efficient training \cite{al2019character}.

\paragraph{Scaling} Based on Observation \ref{train_large}, we successively increase the compound coefficient to scale three systems to larger sizes than all previously trained systems, but train for less steps. For our smallest system, we train for 5 epochs on both the long and the short sequences.\footnote{Since validation loss on the long sequences did not further decrease after 3 epochs, the two larger systems were only trained for 3 epochs on these sequences (cf. Appendix \ref{a:val_loss}).} The results are listed in Table \ref{tab:final_scaling}. Furthermore, Table \ref{tab:verification} shows a comparison of the smallest of the three systems to the best performing system so far, as well as to a modification of this system which fulfills the requirement $A \approx H/64$. As can be observed, both the performance on the large GLUE-Large tasks and the final validation loss are improved, while requiring less training time. For the two larger systems, each obtained by approximately doubling the model size, downstream performance and validation loss are further improved (cf. Tab. \ref{tab:final_scaling}). Note that these systems are rather large compared to the amount of pre-training data. This demonstrates the remarkable robustness of these systems with respect to overfitting on the pre-training data, which is in line with the results of \citet{kaplan2020scaling}.

\section{Conclusion \& Future work}
\label{sec:conc}

\paragraph{Limitations}

The most severe limitation is the small pre-training data set. Based on the observations of \citet{kaplan2020scaling}, systems train faster if more training examples are used. The small size of the pre-training data set might also be the cause of overfitting on smaller tasks. Therefore, for further experiments, we suggest to expand the amount of pre-training data. Furthermore, we did no hyperparameter tuning, but instead adopted the configurations from the original models. It would be advisable to adjust the hyperparameters accordingly \cite{li2020train}, especially since we used different batch sizes as the original models.

\paragraph{Directions for Further Research}

\citet{kaplan2020scaling} studied the effect of the amount of pre-training data, however, not with regard to downstream task performance. Due to the fact that current NLP systems are trained on vastly different amounts of pre-training data, we believe that this relationship should be explored further. 

Although attempts have been made to study the relationship between different pre-training objectives and the performance on downstream tasks \cite{arora2019theoretical}, this relation is yet not well understood. Empirically, contrastive pre-training objectives, such as replaced token detection \citep{clark2020electra} have shown very promising results. It would be interesting to extend the study to such contrastive objectives. Since we observed that the NSP task is beneficial for learning sentence-pair relationships, comparing it to ALBERT's SOP task \citep{lan2019albert} could yield further insights.

Finally, by fine-tuning on a larger variety of tasks we could break down in more detail how different modeling choices affect the performances on different tasks. We believe that further investigation of such relationships will open many opportunities for future research.

\paragraph{Conclusion}

In our experiments, BERT-style systems consistently outperform RoBERTa-style and GPT-2-style systems. We therefore conclude that, at least in case of a relatively small pre-training data set, the combination of MLM \& NSP is preferable to MLM or LM. Although our experiments were conducted on a much smaller scale than other studies, we were able to reproduce many previous findings. For instance, we observed that, provided multiple dimensions are scaled, systems with very different shapes can achieve similar performances.

Consistent with previous studies \citep{kaplan2020scaling,li2020train} we found that it is in general inefficient to train until convergence and that training for more steps improves the performance rather marginally. Instead, in accordance with \citet{kaplan2020scaling}, we believe that increasing the batch size is more beneficial than training for more steps.

More importantly, also consistent with the results of \citet{kaplan2020scaling} and \citet{li2020train}, we conclude that the model size is the key factor in Transformer-based systems. We observed that even for rather large systems, both the final pre-training validation loss and the GLUE performance benefit from further increasing the size. At the same time, the total pre-training time increases at a rather low rate. In particular, given a fixed time budget, large systems reach a lower loss than small systems. Therefore, we believe that additional compute should be allocated mainly to increase the model size.

\clearpage

\bibliographystyle{acl_natbib}
\bibliography{konvens2021}

\clearpage

\section*{Appendix}

\appendix

\section{Pre-training details}\label{a:pretrain}

\paragraph{Training duration} To ensure a fair comparison of the different pre-training objectives, we pre-train RoBERTa-style and GPT-2-style systems for $10$ epochs, and BERT-style systems for $6$ epochs, which in all cases equates to approximately $137,000$ total training steps combined over both partitions.\footnote{In sections where we do not compare the different objectives the number of epochs may differ.} Since the data is duplicated when training with MLM \& NSP, it is natural to simply lower the number of epochs in relation to the amount of pre-training data. While the amount of pre-training data of RoBERTa-style and GPT-2-style systems amounts to more than $60\%$ of the data of BERT-style systems, we found that, on the other hand, the average WordPiece token contains slightly more information than the average byte-level BPE token.

\paragraph{Optimization} Apart from the experiments in section \ref{sec:batchsize_steps}, we use a batch size of $64$ when training on the short sequences and a batch size of $16$ for the long sequences. We optimize all systems with Adam \cite{kingma2014adam} using the following parameters: $\beta_1 = 0.9$, $\beta_2 = 0.999$, $\epsilon = 1\mbox{e}$-$6$ and $L_2$ weight decay of $0.01$. For BERT-style and RoBERTa-style systems we use a maximum learning rate of $1\mbox{e}$-$4$, and for GPT2-style systems the maximum learning rate is $2.5\mbox{e}$-$4$. In all cases we use a linear warmup for the first 1000 steps, which corresponds to approximately $1\%$ of the total steps. Furthermore, for all systems we employ dropout with a rate of $0.1$ on all layers. The activation function of all systems is the GELU \cite{hendrycks2016bridging}. The hyperparameters are in general chosen as in the original systems, except for RoBERTa-style systems, because RoBERTa was trained with significantly larger batches, which requires different hyperparameters. For RoBERTa-style systems we therefore choose the same hyperparameters as for BERT-style systems.

\paragraph{Implementation} We pre-train all systems on a single NVIDIA 16GB V100 GPU, making use of the Hugging Face transformers library \cite{wolf2020transformers}. The same also holds true for fine-tuning.\\

\paragraph{Short and long sequences}

With our pre-training procedure we follow \citet{devlin2019bert}: The first 90\% of the steps on short sequences (128 tokens), the remaining 10\% on long ones (512 tokens). When inspecting the validation loss, we adjust the evaluation sequence lengths to the lengths of the training sequences, so ensure the same distribution for training and validation data. This causes the validation loss on the long sequences to start at a slightly higher point than the final validation loss on the short sequences (cf. Appendix \ref{a:val_loss}).

\section{Fine-tuning details}\label{a:finetune}

We follow \citet{devlin2019bert} and train for three epochs on all GLUE tasks. We use a batch size of $16$ and a learning rate of $2$e-$5$ for each task. Apart from these hyperparameter configurations, we apply the same fine-tuning procedures that were used by the original systems. For GPT-2-style systems, we implemented the fine-tuning approach of GPT (because GPT-2 was not fine-tuned).

However, we do make one small modification to the original implementations. In contrast to BERT-style systems, the pre-training objective of RoBERTa-style and GPT-2-style systems does not contain a classification task. When performing the NSP task, in the original BERT the contextualized representation of the $\texttt{CLS}$ token is obtained by feeding the corresponding final hidden state through a linear layer with dropout and $tanh$ activation. Subsequently, the contextualized representation is fed through another linear layer with dropout, which is the output layer mapping the contextualized representation to the class probabilities. Consequently, when fine-tuning BERT-style systems on a classification task, there are in fact two linear layers between the final hidden state and the output classes. However, RoBERTa and GPT in their original implementation use only one linear layer. In order to be as consistent as possible, in contrast, we use two linear output layers for all systems. The first linear layer is followed by a $tanh$ activation and both layers are implemented with a dropout rate of $0.1$. For more information regarding this issue see \href{https://discuss.huggingface.co/t/what-is-the-purpose-of-the-additional-dense-layer-in-classification-heads/526}{huggingface's discussion forum}.

\section{Detailed performance values for single shape dimensions and results for the grid search}\label{a:performance}

Performance values on GLUE-Large and SST-2 for scaling $H$ (Tab. \ref{tab:embed}) and for scaling $L$ (Tab. \ref{tab:layer}). Table \ref{tab:grid_search} shows the results of the grid search.

\begin{table*}[ht]
\centering
\begin{adjustbox}{width=.8\textwidth}
\renewcommand{\arraystretch}{0.8}
\begin{tabular*}{\textwidth}{@{\extracolsep{\fill}} ccccccccc}
\btrule{1.5pt}
\multicolumn{4}{c}{\textbf{BERT-Style}} &
\multicolumn{5}{c}{\textbf{Validation Set Performance}} \\
\cmidrule(r){1-4}
\cmidrule(r){5-9}
A & H & L & $N_{\text{model}}$ & $\text{GLUE-Large}$ & MNLI-(m/mm) & QQP &  QNLI & SST-2 \\
\midrule
2 & 128 & 2 & 393,216 & 65.4 & $59.0/60.2$ & $72.3$ & $64.8$ & $78.0$ \\
2 & 192 & 2 & 884,736 & 67.2 & $62.1/62.8$  & $74.0$  & $65.4$ & $82.6$ \\
2 & 288 & 2 & 1,990,656 & 69.3 & $63.7/65.2$ & $76.0$ & $68.3$ & $82.0$\\
2 & 384 & 2 & 3,538,944 & 72.3 & $65.7/66.6$ &  $77.8$ & $73.2$ & $81.1$\\
2 & 544 &  2 & 7,102,464 & 72.3 & $66.8/68.1$ & $78.0$ & $72.0$ & $83.3$ \\
\midrule
\multicolumn{4}{c}{\textbf{GPT-2-Style}} &
\multicolumn{5}{c}{\textbf{Validation Set Performance}} \\
\cmidrule(r){1-4}
\cmidrule(r){5-9}
A & H & L & $N_{\text{model}}$ & $\text{GLUE-Large}$ & MNLI-(m/mm) & QQP &  QNLI & SST-2\\
\midrule 
2 & 128 & 2 & 393,216 & 61.6 & $56.3/56.2$ & $66.1$ & $62.3$ & $79.8$ \\
2 & 192 & 2 & 884,736 & 62.9 & $58.0/58.4$  & $68.7$  & $61.9$ & $79.7$\\
2 & 288 & 2 & 1,990,656 & 63.9 & $58.7/58.7$ & $70.9$ & $62.2$ & $81.7$\\
2 & 384 & 2 & 3,538,944 & 64.9 & $59.8/59.6$ &  $71.9$ & $63.0$ & $81.2$\\
2 & 544 & 2 & 7,102,464 & 65.0 & $59.8/59.7$ & $72.4$ & $62.9$ & $82.5$\\
\midrule
\multicolumn{4}{c}{\textbf{RoBERTa-Style}} &
\multicolumn{5}{c}{\textbf{Validation Set Performance}} \\
\cmidrule(r){1-4}
\cmidrule(r){5-9}
A & H & L & $N_{\text{model}}$ & $\text{GLUE-Large}$ & MNLI-(m/mm) & QQP &  QNLI & SST-2\\
\midrule
2 & 128 & 2 & 393,216 & 60.1 & $53.7/55.1$ & $64.7$ & $61.9$ & $79.2$ \\
2 & 192 & 2 & 884,736 & 60.5 & $54.4/55.4$  & $65.0$  & $62.0$ & $80.8$ \\
2 & 288 & 2 & 1,990,656 & 63.0 & $57.5/58.0$ & $68.1$ & $63.4$ & $80.3$ \\
2 & 384 & 2 & 3,538,944 & 64.3 & $59.4/59.8$ &  $69.0$ & $64.6$ & $81.9$\\
2 & 544 &  2 & 7,102,464 & 66.5 & $60.2/60.7$ & $72.7$ & $66.5$ & $81.8$\\
\btrule{1.5pt}
\end{tabular*}
\end{adjustbox}
\caption{Performance on GLUE when increasing only the embedding dimension.}
\label{tab:embed}
\end{table*}

\begin{table*}[ht]
\centering
\begin{adjustbox}{width=.8\textwidth}
\renewcommand{\arraystretch}{0.8}
\begin{tabular*}{\textwidth}{ @{\extracolsep{\fill}} ccccccccc}
\btrule{1.5pt}
\multicolumn{4}{c}{\textbf{BERT-Style}} &
\multicolumn{5}{c}{\textbf{Validation Set Performance}} \\
\cmidrule(r){1-4}
\cmidrule(r){5-9}
A & H & L & $N_{\text{model}}$ & $\text{GLUE-Large}$ & MNLI-(m/mm) & QQP &  QNLI & SST-2 \\
\midrule
2 & 128 & 2 & 393,216 & 65.4 & $59.0/60.2$ & $72.3$ & $64.8$ & $78.0$ \\
2 & 128 & 5 & 983,040 & 68.9 & $62.1/64.2$  & $75.0$  & $68.6$ & $79.8$ \\
2 & 128 & 10 & 1,966,080 & 72.0 & $65.3/66.9$ & $76.7$ & $74.1$ & $81.8$ \\
2 & 128 & 18 & 3,538,944 & 74.2 & $67.2/68.6$ &  $77.8$ & $77.7$ & $82.2$ \\
2 & 128 &  36 & 7,077,888 & 75.9 & $69.7/70.4$ & $79.7$ & $78.3$ & $83.3$ \\
\midrule
\multicolumn{4}{c}{\textbf{GPT-2-Style}} &
\multicolumn{5}{c}{\textbf{Validation Set Performance}} \\
\cmidrule(r){1-4}
\cmidrule(r){5-9}
A & H & L & $N_{\text{model}}$ & $\text{GLUE-Large}$ & MNLI-(m/mm) & QQP &  QNLI & SST-2\\
\midrule 
2 & 128 & 2 & 393,216 & 61.6 & $56.3/56.2$ & $66.1$ & $62.3$ & $79.8$ \\
2 & 128 & 5 & 983,040 & 62.4 & $57.6/56.1$  & $67.4$  & $62.0$ & $80.5$ \\
2 & 128 & 10 & 1,966,080 & 62.0 & $56.9/57.0$ & $67.7$ & $61.5$ & $81.4$ \\
2 & 128 & 18 & 3,538,944 & 61.8 & $56.1/56.4$ &  $66.8$ & $62.4$ & $80.6$ \\
2 & 128 &  36 & 7,077,888 &61.4 & $56.6/56.7$ & $66.6$ & $61.1$ & $80.7$ \\
\midrule
\multicolumn{4}{c}{\textbf{RoBERTa-Style}} &
\multicolumn{5}{c}{\textbf{Validation Set Performance}} \\
\cmidrule(r){1-4}
\cmidrule(r){5-9}
A & H & L & $N_{\text{model}}$ & $\text{GLUE-Large}$ & MNLI-(m/mm) & QQP &  QNLI & SST-2\\
\midrule
2 & 128 & 2 & 393,216 & 60.1 & $53.7/55.1$ & $64.7$ & $61.9$ & $79.2$ \\
2 & 128 & 5 & 983,040 & 64.8 & $59.5/60.6$  & $70.4$  & $64.4$ & $80.2$ \\
2 & 128 & 10 & 1,966,080 & 67.1 & $60.9/61.9$ & $72.0$ & $68.5$ & $81.7$\\
2 & 128 & 18 & 3,538,944 & 67.2 & $62.9/64.3$ &  $74.3$ & $64.3$ & $80.0$\\
2 & 128 &  36 & 7,077,888 & 73.3 & $67.6/69.1$ & $77.3$ & $75.0$ & $82.6$\\
\btrule{1.5pt}
\end{tabular*}
\end{adjustbox}
\caption{Performance on GLUE when increasing only the number of layers.}
\label{tab:layer} 
\end{table*}

\begin{table*}[ht]
\centering
\begin{adjustbox}{width=.8\textwidth}
\renewcommand{\arraystretch}{0.8}
\begin{tabular*}{\textwidth}{ @{\extracolsep{\fill}} cccccc}
\btrule{1.5pt}
\multicolumn{4}{c}{\textbf{BERT-Style}} &
\textbf{Validation Loss (WikiText-103)} &
\textbf{Validation Performance (GLUE)} \\
\cmidrule(r){1-4}
\cmidrule(r){5-5}
\cmidrule(r){6-6}
A & H & L & $N_{\text{model}}$ & BERT-Style Loss & $\text{GLUE-Large}$ \\
\midrule
2 & 128 & 2 & 393,216 & \textbf{5.66} & 66.6 \\
2 & 104 & 3 & 389,376 & \textbf{6.34} & \textbf{68.2} \\
2 & 90 &4 & 388,800 & \textbf{6.41} & 67.1 \\
2 & 74 & 6 & 394,272 & 6.47 & - \\
2 & 64 & 8 & 393,216 & 6.50 & - \\
2 & 58 & 10 & 403,680 & 6.54 & - \\
2 & 52 & 12 & 389,376 & 6.58 & - \\
2 & 48 & 14 & 387,072 & 6.62 & - \\
2 & 46 & 16 & 406,272 & 6.62 & - \\
\btrule{1.5pt}
\end{tabular*}
\end{adjustbox}
\caption{Grid search over nine small BERT-style systems.}
\label{tab:grid_search} 
\end{table*}

\clearpage

\section{Validation loss for scaled-up models}\label{a:val_loss}

\vspace{-3.25cm}

\begin{figure}[ht]
  \centering
  \captionsetup{margin=0cm}
  \begin{subfigure}[b]{1\linewidth}
  \includegraphics[width=\linewidth]{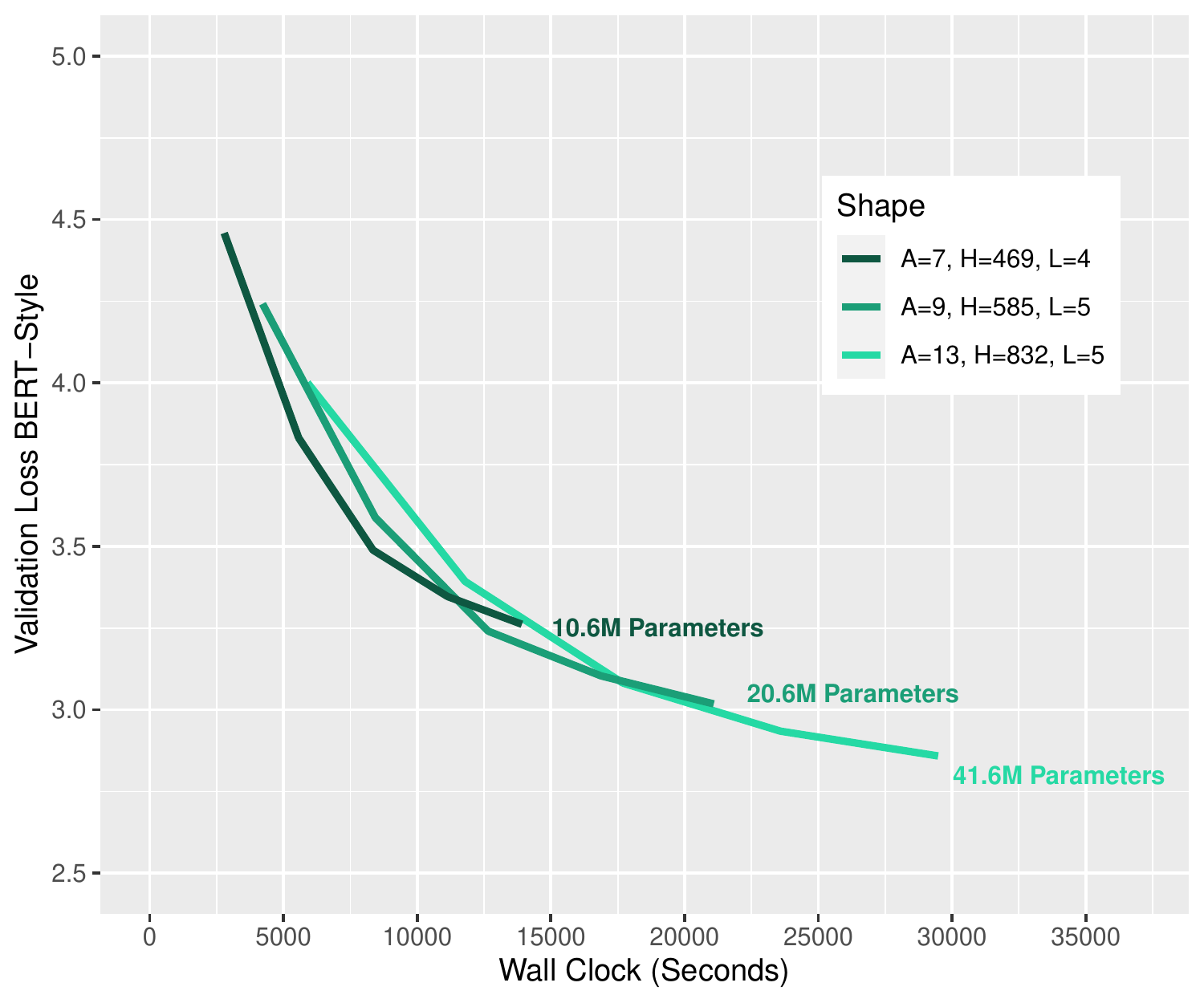}
  \end{subfigure}
  \caption{Validation loss of scaled-up BERT-style systems when pre-training on the short sequences. The depicted parameter counts refer to $N_{model}$.}
  \label{fig:loss_short}
\end{figure}

\vspace{-4cm}

\section{Definition of the model size}\label{a:modelsize}

\vspace{-2cm}

We follow \citet{kaplan2020scaling} and use the approximate number of non-embedding parameters to define the model size, which we denote as $N_{\text{model}}$. The embedding parameters consist of all token, position and (if present) segment embeddings. The number of embedding parameters does not depend on the network depth, and when scaling width and/or depth, it is a sub-leading term of the total number of parameters. Furthermore, the number of FLOPs related to embedding (and de-embedding) is also sub-leading term of the total number of FLOPs. Consistent with this is the observation of \citet{kaplan2020scaling} that discarding the number of embedding parameters when calculating model size and amount of compute results in significantly cleaner scaling laws. Since the share of embedding parameters decreases significantly for larger models, similarly to \citet{kaplan2020scaling} we expect that discarding the number of embedding parameters allows for a better generalization of our results to large models. Another advantage of defining the model size as the number of non-embedding parameters is that this number is closely linked to the number of (non-embedding related) FLOPs. This enables us to design benchmarking scenarios by training different models of comparable size, which at the same time require roughly similar amounts of computation. 

\begin{figure}[ht]
  \centering
  \captionsetup{margin=0cm}
  \begin{subfigure}[b]{1\linewidth}
  \includegraphics[width=\linewidth]{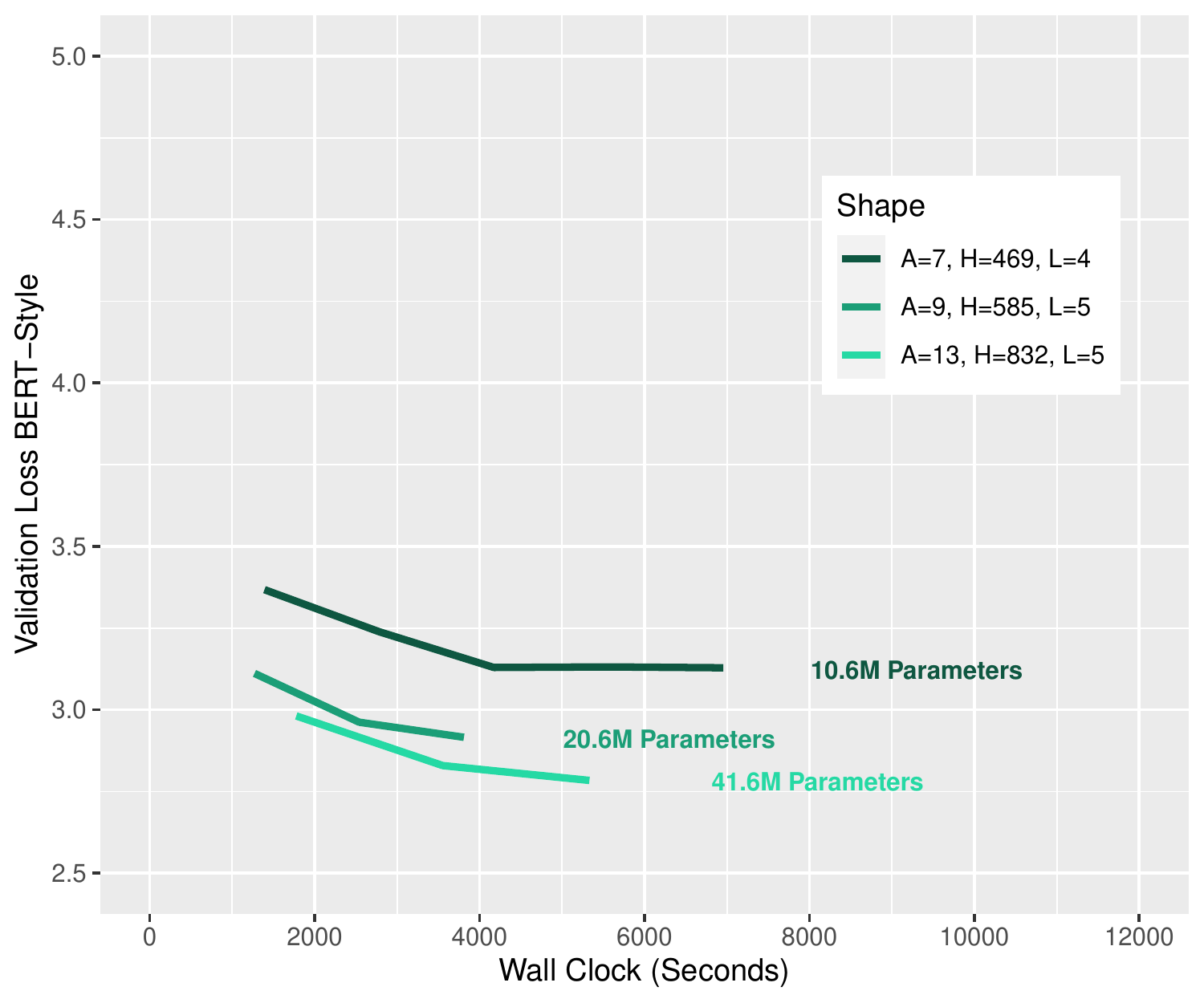}
  \end{subfigure}
  \caption{Validation loss of scaled-up BERT-style systems when pre-training on the long sequences. The depicted parameter counts refer to $N_{model}$.}
  \label{fig:loss_long}
\end{figure}

\subsection*{Number of Non-Embedding Parameters}

\noindent Omitting biases and other sub-leading terms, the number of non-embedding parameters, which is our definition of the model size, is given by
\begin{align}
N_{\text{model}} := 12 L H^2,
\end{align}
where we have assumed that $H_k = H_v = \frac{H}{A}$ and $H_{ff} = 4H$. Therefore, per layer there are approximately $12H^2$ non-embedding parameters. This number can be derived from the following three steps performed in each layer of a Transformer:

\paragraph{1. Input projection} For each attention head, the queries, keys and values of dimension $\frac{H}{A}$ are obtained with the three matrices $\boldsymbol{W}_i^Q$, $\boldsymbol{W}_i^K$, and $\boldsymbol{W}_i^V$, which are each of size $H \times \frac{H}{A}$. In total, the input projection thus consists of $3 \cdot A \cdot \frac{H^2}{A} = 3 H^2$ parameters.
\paragraph{2. Output projection} First, note that performing attention on the projected inputs of dimension $\frac{H}{A}$ involves no additional parameters. The concatenated attention results are projected back to dimension $H$ with the $H \times H$ matrix $\boldsymbol{W}^O$. Therefore, the output projection involves an additional set of $H^2$ parameters.
\paragraph{3. Feed-forward network} The last sub-layer of each layer consists of applying a feed-forward network to the output projections. There exist $H \cdot 4H$ connections between the output projections and the neurons of the inner-layer, and another $4H \cdot H$ connections from the inner-layer to the final output neurons. This step hence involves $8H^2$ parameters.

\noindent Note that the feed-forward network accounts for the majority of non-embedding parameters, followed by the input and output projections, respectively.

\subsection*{Relation to FLOPs}

As stated, the number of non-embedding parameters is closely linked to the number of non-embedding related FLOPs. We start by deriving the number of FLOPs per token and forward pass for GPT-2-style systems, where sub-leading terms such as biases and layer normalization are again omitted.

\paragraph{1. Input projection} The matrix-vector products of each per-layer input with $\boldsymbol{W}_i^Q$, $\boldsymbol{W}_i^K$, and $\boldsymbol{W}_i^V$ involve approximately $3 \cdot 2 \cdot H \cdot \frac{H}{A}$ FLOPs per attention head. Considering all attention heads, the input projection thus requires approximately $6 H^2$ FLOPs per token.
\paragraph{2. Attention} The computation of the attention operation can be divided into two sub-components:
\begin{itemize}
\item \textbf{Computation of the weights}: On average, $\frac{N_{ctx}}{2}$ attention weights have to be computed per input token, since on average half of the tokens are masked for each input token. Computation of a dot-product attention weight requires approximately $2 \frac{H}{A}$ FLOPs per head. In total, the computation of the attention weights hence involves approximately $N_{ctx} H$ FLOPs per token.
\item \textbf{Computation of the weighted sum}: Since only half of the tokens are summed on average, given the attention weights, calculation of the weighted sum of the values has an average cost of approximately  $N_{ctx} H$ FLOPs for each token.
\end{itemize}
\paragraph{3. Output projection} The vector matrix product of the attention outputs with $\boldsymbol{W}^O$ requires approximately $2 H^2$ FLOPs for each token.
\paragraph{4. Feed-forward network} The feed-forward network consists of two consecutive matrix multiplications, where each matrix contains $4H^2$ parameters. Thus, the feed-forward network requires approximately $2 \cdot 2 \cdot 4H^2 = 16 H^2$ FLOPs per token. 

The number of FLOPs per token and forward pass in GPT-2-style systems, which we denote by $C_{forward}$, can hence be approximated as
\begin{align}
\begin{split}
C_{forward} & \approx L(6 H^2 + N_{ctx} H + N_{ctx} H \\
            & \quad + 2 H^2 + 16 H^2) \\
& = 24 LH^2 + 2 L N_{ctx} H \\
& = 2N_{\text{model}} + 2 L N_{ctx} H.
\end{split}
\label{forward_pass_gpt}
\end{align}
BERT-style and RoBERTa-style systems require slightly more FLOPs than GPT-2-style systems, because these systems have no autoregressive attention mask. Hence, in both steps of the attention operation above, the computational cost is approximately twice as much, i.e., $2N_{ctx} H$ in each step. Therefore, BERT-style and RoBERTa-style systems require approximately $2N_{\text{model}} + 4 L N_{ctx} H$ FLOPs per token and forward pass. As mentioned by \citet{kaplan2020scaling}, if $H > N_{ctx}/12$, the context-dependent term in Eq. \eqref{forward_pass_gpt} only accounts for a relatively small fraction of the compute of GPT-2-style systems. In particular, when increasing $H$, the importance of the context-dependent term diminishes. For BERT-style and RoBERTa-style systems the context-dependent term becomes small if $H > N_{ctx}/6$. Both constraints are satisfied by a large margin for all our systems, especially since we mainly train on rather short sequences. The backward pass requires approximately twice as much compute as the forward pass \cite{kaplan2020scaling}, such that the total amount of non-embedding related compute per token and training step can be approximated as 
\begin{align}
C := 6N_{\text{model}}.
\end{align}

\section{Sequence characteristics}\label{a:input}

The following Table \ref{tab:input} provides an overview on the number of tokens in short and long sequences.

\begin{table}[ht]
\begin{adjustbox}{width=.49\textwidth}
\begin{tabular}{cccc}
\btrule{1.5pt}
\textbf{System} & \textbf{Partition} & \multicolumn{2}{c}{\textbf{Number of Tokens}} \\
  &   & Total & Average \\
\cmidrule(r){1-1}
\cmidrule(r){2-2}
\cmidrule(r){3-4}
BERT-Style & Short & $110,888,186$ & $110.04$ \\
           & Long & $43,274,856$  & $375.52$ \\
                           
RoBERTa-Style & Short & $70,025,709$   & $110.31$\\
              & Long & $27,692,351$   &  $457.04$\\

GPT-2-Style  &  Short & $70,564,106$   & $111.16$\\
             & Long & $27,729,551$   & $457.65$\\
\btrule{1.5pt}
\end{tabular}
\end{adjustbox}
\caption{Number of tokens for the short and the long sequences as well as the average sequence lengths resulting from the different tokenizers.}
\label{tab:input}
\end{table}

\end{document}